\documentclass[runningheads]{llncs}
\usepackage{graphicx}
\usepackage{microtype}
\usepackage{xcolor}
\usepackage{wrapfig}
\usepackage{subfigure}
\usepackage{pgfplots}
\usepackage{booktabs} 
\PassOptionsToPackage{hyphens}{url}\usepackage{hyperref}
\usepackage{textcomp} 

\usepackage{amsmath}
\DeclareMathOperator*{\argmax}{argmax}
\usepackage{bm}
\usepackage{float}
\begin{document}
\title{Evaluating Tree Explanation Methods for Anomaly Reasoning: A Case Study of SHAP TreeExplainer and TreeInterpreter}
\titlerunning{Evaluating Tree Explanation Methods for Anomaly Reasoning}
%
\author{Pulkit Sharma\inst{1} \and
Shezan Rohinton Mirzan\inst{1} \and
Apurva Bhandari\inst{1} \and 
Anish Pimpley\inst{2} \and 
Abhiram Eswaran\inst{2} \and 
Soundar Srinivasan\inst{2} \and
Liqun Shao\inst{2}}

 \authorrunning{P. Sharma et al.}

\institute{University of Massachusetts, Amherst MA 01002, USA\\
\email{\{psharma, smirzan, apurvabhanda\}@umass.edu}
\and
Microsoft Corp., Cambridge MA, 02142, USA\\
\email{\{anpimple, abeswara, sosrini, lishao\}@microsoft.com}}
\maketitle              
\begin{abstract}
Understanding predictions made by Machine Learning models is critical in many applications. 
In this work, we investigate the performance of two methods for explaining tree-based models: `Tree Interpreter (TI)' and `SHapley Additive exPlanations TreeExplainer (SHAP-TE)'. Using a case study on detecting anomalies in job runtimes of applications that utilize cloud-computing platforms, we compare these approaches using a variety of metrics, including computation time, significance of attribution value, and explanation accuracy. We find that, although the SHAP-TE offers consistency guarantees over TI, at the cost of increased computation, consistency does not necessarily improve the explanation performance in our case study.

\end{abstract}
\section{Introduction}
\label{introduction}

Machine learning-based approaches have become popular in automatically detecting and predicting anomalies in a variety of applications. 
   Anomaly Detection (AD) \cite{anomDetSurvey} has been applied to various domains, such as tracking anomalous events in traffic surveillance videos \cite{SultaniCVPR} and tracking irregular patterns in electrocardiographs of a patient \cite{anomalyEEG}. 
    AD has also become popular in the computing industry to detect system failures both in multi-server distributed systems \cite{padmulti} and in embedded systems\cite{runtimeanomaly}. A sister domain closely related to AD is Anomaly Reasoning (AR) that comprises delineating the causal factors associated with an anomaly. A robust system would not just predict the anomalous events, but also identify the root causes of a failure or an anomaly. In these applications, AR is crucial for an efficient analysis of faults, thereby reducing significantly, the time needed for manual investigation and the required computing resources.

     Interpreting machine learning models correctly has been a challenging task. Although linear models are easy to interpret, they fail to generalize in real-world scenarios with predominantly non-linear behavior. This leads to the adoption of more accurate models at the cost of them being less interpretable. Hence, there has been a significant recent research emphasis on developing techniques that could add interpretability to these complex `black-box models'. 

    The~\textit{LIME} \cite{lime} algorithm interprets the predictions of any given model by utilizing explainable analogues that are valid in ``local" regions. Another popular apporach, \textit{DeepLIFT} ~\cite{deeplift17,deeplift16}, is used for interpreting Neural Network-based models by assigning contribution scores to each neuron based on the difference in its activation to its "reference" activation. Other techniques~\cite{shap,shapleySampling,shapleyRegression} take a game-theoretic approach towards computing feature contributions to model explanations. Although,  all three are primarily based on Shapley values \cite{kuhn1953contributions}, they  calculate and further approximate the values differently to derive feature contributions.

  In this paper, we compare the interpretation performance of two popular tree-explanation methods: the SHapley Additive exPlanation TreeExplainer (SHAP-TE) \cite{treeExplainer} for model-agnostic interpretations and the TreeInterpreter (TI) \cite{treeInterpreterGit}.
  Specifically, we conduct a case study on the task of reasoning about anomalies in computing jobs that run in cloud platforms. An example of a recent effort in this domain is Griffon \cite{Shao_2019} - Microsoft's Reasoning infrastructure deployed on Azure clusters. 
  SHAP-TE averages out contributions of each possible feature set to obtain the final feature attribution values. This helps in reducing the bias added to the computation of feature attribution values when only a specific ordering is considered. However, TI only considers one ordering of features depending on how the tree was formed. This makes TreeInterpreter as essentially an approximation of SHAP-TE. Hence, we examine the performance of SHAP-TE using Griffon's AD algorithm and investigate if the TreeInterpreter's approximation-based approach used in Griffon could be generalized to other datasets or scenarios. 
  
  We conduct experiments under a variety of conditions on the recently introduced PostgreSQL dataset \cite{NIPS2019_9345} and compare the above methods across a variety of metrics.
    Our major contributions are summarized as follows:
    
    \begin{itemize}
        \item \textbf{Scale Comparison.} We compare, empirically, the scaling property of both methods, in terms of time-complexity, with respect to increasing data size and depth of trees.
        \item \textbf{Performance Comparison.} We analyze their trade-offs and provide a novel analysis of the two methods in ranking features according to their contributions and attribution accuracy, and also experimentally evaluate the variance among the contribution values generated by the two methods to measure the significance of produced ordering of FAs\footnote{\textit{Feature Attribution} (FA) is defined as the contribution each independent variable or a ``feature" made to the final prediction of a model.}.
        \item \textbf{Critique of TI and SHAP-TE}. We investigate whether the \textit{consistency~\footnote{\textit{See Section~\ref{relatedWork} for the definition of consistency.}}} property of SHAP-TE is crucial to  making it a preferred method over TI in this domain on publicly available data, which would facilitate replication by the research community.
    \end{itemize}
\section{Background Work}
\label{relatedWork}
Unlike linear models, Decision Tree models cannot be represented as sum of linear contributions of each features for the whole model. Hence, they are difficult to interpret on the model level. 
However, individual predictions of a decision tree can be explained by decomposing the decision path into one component per feature. One can track a decision by traversing the tree and explain a prediction ($y$) by the additive contributions at each decision node as,
\begin{equation}
    y = \mathrm{bias} + \sum_{m=1}^{M} \mathrm{feature\_contribution}\{m,x\}\label{ti}
\end{equation}
where $\mathrm{bias}$ is the contribution of root node and $\mathrm{feature\_contribution}\{m,x\}$ is the contribution of feature $m$ in predicting the outcome corresponding to an input $x$. 
Equation~(\ref{ti}) forms the basis of TI Package ~\cite{treeInterpreterGit}, which is available for interpreting scikit-learn's decision tree and random forest predictions.

SHAP-TE ~\cite{treeExplainer} is another tree-based FAM\footnote{\textit{Feature Attribution Method} (FAM), referred to as the explanation method that calculates FAs to interpret each prediction generated by a model.} that uses Shapley values from game theory to make tree-based models interpretable. The feature values of an input data instance act as players in a coalition. These values essentially distribute the prediction result among different features. SHapley Additive exPlanation (SHAP) \cite{shap} value for a particular feature is the weighted average of all the marginal contributions to the prediction over all possible combinations of features. 
Using SHAP is inherently beneficial in terms of \textit{consistency}. From \cite{shap}, a FAM is consistent if a feature's attribution value does not decrease on increasing the true impact of that feature in the model.

Griffon\cite{Shao_2019}, introduced in Section \ref{introduction}, uses the notion of~\textbf{delta feature contribution} between two different instances of a job type to predict features that impacted the most to the difference in their runtimes. FA values for both the jobs are simply subtracted to compute how much each feature contributes to the deviation in the predictions for both jobs.

 
Work on the evaluation of Causal Models in~\cite{NIPS2019_9345} discusses the design of interventional and observational data to analyze the performance of model explanation paradigms. Motivated by their approach, we also adopt the postgreSQL data presented in their work in the experiments presented in this paper. This dataset is a collection of SQL queries submitted to stackoverflow server's database (large-scale software system) that enables the experimenter to run the same experiment multiple times under different, controlled conditions. The conditions are determined by setting few key configuration parameters, called \textbf{treatment} variables. The effect of change in one configuration is observed in the \textbf{output} variables such as Runtime. 
Using background, domain knowledge, the dependence between the runtime and the "treatment" variables is established. The features that describe a job are termed as \textbf{covariates}. Inspired by this work, we adopt the interventional data setting within the scope of Griffon; design of these experiments is discussed in Section \ref{experiments}.

\section{Evaluation Approach}
\label{evaluation_approach}
In order to group similar data points together, we split the entire data set into smaller subgroups termed as \textit{templates}. A template is a group of data points in which the \textit{covariate} input features are kept nearly constant\footnote{Some of the covariate variables in postgreSQL dataset are continuous, which when grouped reduces the number of data points per cluster.} while the \textit{treatment} feature variables are allowed to change for each data point. This approach allows us to easily establish a relationship between the prediction and the treatment feature variable for any two datapoints belonging to the same template. Hence, if we choose a pair of datapoints from a template such that both differ with respect to some $m$ treatment feature variables, then the deviation in their predictions can be easily attributed to these $m$ features.

\begin{figure}
    \centering
    \includegraphics[scale=0.4]{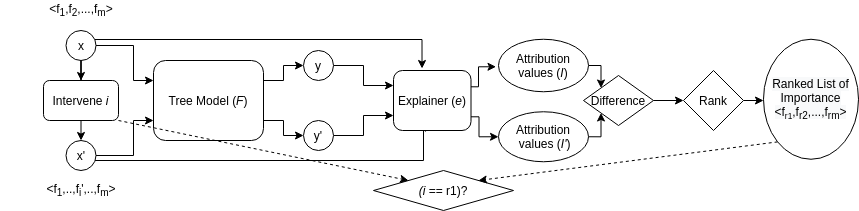}
    \caption{Overview of the proposed evaluation approach. ($\mathbf{x}$, $\mathbf{y}$) and ($\mathbf{x'}$, $\mathbf{y'}$) are the input, output pairs for original and intervened data points, respectively. Explainer ($\mathbf{e}$) refers to one of the explanation methods considered in our work and produces the FA values corresponding to a prediction. A ranked list is obtained based on their differences and then check whether the top attributed feature is the same as the intervened one.}
    \label{fig:FlowChart}
\end{figure}

For the case of anomaly detection, we consider pairs of data points from a particular template. FAM will calculate FA values for both these data points. These obtained FA values can then be used to reason about the output. In order to explain the difference in their predictions, we compute the difference in the computed FA values (See Figure \ref{fig:FlowChart}). This is motivated from the delta feature contribution technique used in Griffon. Using the obtained delta feature contributions, we create an ordered list that ranks all the features from the most important to the least important feature.
From this list of ranked features, we compute a Rank Biased overlap (RBO) metric \cite{rbo} \footnote{RBO implementation :  \href{https://github.com/changyaochen/rbo.}{https://github.com/changyaochen/rbo}} to measure the similarity of ranked results produced by both the attribution methods. 
Motivated by the differences in RBO values observed \ref{tab:rbo}, we further investigate the differences brought out by lower RBO values in terms of \textit{attribution accuracy}. We define attribution accuracy as the fraction of times a FAM is able to correctly identify the cause of an anomaly. To assess the trade-off between using SHAP-TE and TI, we measure the attribution accuracy on interventional PostgreSQL data set using proposed Implicit and Explicit Interventional Measures. 


\subsection{Implicit Interventional Measure}
\label{interventional_measure_approach}

We use the ``Interventional Data" setup proposed by \cite{NIPS2019_9345} to establish the attribution accuracy of a particular method of explanation. We refer to this approach as an ``Implicit Interventional approach" due to the presence of pre-existing treatment variables in our evaluation data and contrast it with a manually manipulated ``Explicit Interventional approach" that we will introduce in Section~\ref{explicit_interventional_measure_approach}.

We first divide the data set into a set of templates and then choose a pair of data points from within each template. In this pair, one point acts as an anomaly, while the other as the base line. We obtain a ranked feature list after processing this pair of points through an explanation method. The attribution is deemed to be correct, if the top attributed feature matches the treatment variable (see Section~\ref{experiments}). This gives us the Top-$1$ measure of attribution accuracy. Similarly by considering the first $k$ elements of the ranked list, we obtain the Top-$k$ attribution accuracy. 

\subsection{Explicit Interventional Measure}
\label{explicit_interventional_measure_approach}
This measure helps us in evaluating correctness of each explanation method irrespective of deployment. We design this experiment by manually intervening in the data set to change the value of a treatment variable. Let $d$ be the dimension of feature space in data set and $i$ represent the index of treatment variable that was intervened. When we change the $i^{th}$ index for a data point $\bm{x}$,  we obtain a new intervened point, $\bm{x}'$ (Shown in Figure \ref{fig:FlowChart}). Let $e(\cdot, F)$ represent the operator to produce FA values for a data point under a trained Random Forest model $F$. The most important feature is then considered to be the one with highest difference in  attribution value between the original and manually-intervened data point. Here, our assumption is that this feature should be the same as the manually- or explicitly-intervened feature $i$. 
\begin{equation}
\centering
    i = \argmax |e(\bm{x}', F) - e(\bm{x}, F)|
\end{equation}
Every data point that satisfies the above relation for a particular explanation method, contributes as a positive sample for that explanation method. For PostgreSQL data set used in our experiments described in Section~\ref{experiments}, we apply the following two interventions on all the data points for all three treatment variables,
\begin{equation}
\centering
    \bm{x_i} \leftarrow (\bm{x_i} \pm 1) \mod 3
\end{equation}

Since the range of each treatment variable is 0-2, the above operation covers the domain of each variable. As discussed earlier, change in the values of treatment variables would affect the run-time of the job as these variables correspond to system settings while running a query. Results for this experiment are discussed in Section \ref{explicit_interventional_measure_discussion}.

\section{Experiments and Results}
\label{experiments}

\subsection{Experiment and Data Settings}
We perform our experiments on the PostgreSQL\footnote{Dataset can be found at \href{https://groups.cs.umass.edu/kdl/causal-eval-data}{https://groups.cs.umass.edu/kdl/causal-eval-data}} data set that is a sample of the data from user generated queries on Stack Overflow\footnote{This data is collected in the work by \cite{NIPS2019_9345}.}. It contains the execution information for $11,252$ queries run corresponding to $90,016$ different covariate-treatment combinations on Postgres. Treatment variables for this data set are three system parameters, namely (i) \textbf{MemoryLevel}: the amount of system memory available, (ii) \textbf{IndexLevel}: the type of indexing used for accessing a query in database and (iii) \textbf{PageCost}: the type of disk page access. Various outputs of each query are recorded. We choose query $Runtime$ output to analyse our results.

The PostgreSQL dataset is a large dataset with nearly $1.5$ million data points. Training a Random Forest model on the complete data generates a large model and this in turn increases the time for computing feature attributions, as explained in Section \ref{RunTime}. To keep computation times tractable, we sample a subset from the entire data set ensuring that these samples preserve most of the unique samples in the original data set. After sampling, we have about $70,000$ data points, which are further sub-sampled into
 $60\%$--$20\%$--$20\%$ \textit{train-val-test} splits. We run our experiments on a cluster system with Xeon E$5$-$2680$ v$4$ @ $2.40$GHz processor and $128$GB of RAM. 

SHAP-TE and TI can be compared using any tree-based models including Decision Trees, GBMs, and Random Forest. However, among these options, Random Forest models have empirically been shown to have higher accuracy, especially for high-dimensional data in most real world scenarios~\cite{empericalRF}. Apart from Griffon, other anomaly detection research~\cite{otherRFExample} ,\cite{anotherRFExample} also use Random Forests as their base model.
Hence, we choose to use Random Forest for comparing results.

\subsection{Runtime Comparison}
\label{RunTime}

We train a Random Forest Regression model to predict job runtime in milliseconds with a hyper-parameter setting of 200 estimators and a maximum depth 20.
As running times are generally log-normally distributed, we use $logarithm$ of Runtime as the output target. 

With large number of jobs in a given computing cluster, the ability to detect and interpret anomalies in real time becomes crucial. 
SHAP-TE and TI both have an amortized computation-time complexity of $\mathcal{O}(TLD^2)$ where $T$ is the number of individual estimators, $L$ is the maximum number of leaves in any tree and $D$ is the maximum depth of any tree. Although the amortized time complexity is the same for both, their practical running times differ significantly.


    

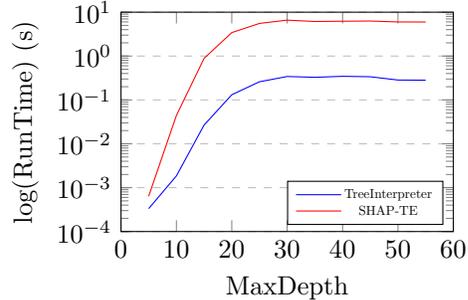
\begin{wrapfigure}[15]{r}{0.5\textwidth}
\label{fig:timing_plots}
\begin{tikzpicture}
\pgfplotsset{width=6cm,height=4.5cm,compat=1.16}
\begin{semilogyaxis}[
    xlabel={MaxDepth},
    ylabel={log(RunTime) (s)},
    xmin=0, xmax=60,
    ymin=0.0001, ymax=10,
    xtick={0,10,20,30,40,50,60},
    ytick={0.0001, 0.001,0.01,0.1,1, 10},
    legend pos=south east,
    ymajorgrids=true,
    grid style=dashed,
    legend style={nodes={scale=0.5, transform shape}}
]

\addplot[color=blue,]
coordinates {
 (5, 0.0003333333)
(10,0.001855555556)
(15,0.02677777778)
(20,0.1313333333)
(25,0.2593333333)
(30,0.3422222222)
(35,0.3277777778)
(40,0.3455555556)
(45,0.3377777778)
(50,0.2841111111)
(55,0.2818888889)
};
\legend{TreeInterpreter}

\addplot[color=red,]
coordinates { (5, 0.00064)
(10, 0.04423)
(15, 0.8870422222)
(20, 3.423933333)
(25, 5.543888889)
(30, 6.577111111)
(35, 6.132222222)
(40, 6.217222222)
(45, 6.283111111)
(50, 5.984)
(55, 5.974111111)
};
\addlegendentry{SHAP-TE}

\end{semilogyaxis}
\end{tikzpicture}
\caption{Comparison of Running Time vs MaxDepth. SHAP-TE scales much faster than TreeInterpreter}
\end{wrapfigure}

First, we compare the scaling of FAM computation time per data instance. We observe that TI scales at $0.058$s per data instance while SHAP-TE takes $3.44$s per data instance which is significantly higher. Second, red lines in Figure \ref{fig:timing_plots} show the scaling of both methods with the depth of tree parameter of the Random Forest. We observe that both methods follow a square dependence, but SHAP-TE again scales worse that TI. The reason is that SHAP-TE averages out the contributions for each decision path while TI just considers the one decision path resulting in shorter computation time. This indicates that in a practical deployment scenario with large number of data cases and larger tree depth, TI can be more efficient compared to the SHAP-TE, in terms of computation times. 

\subsection{Rank List Similarity}
We now report results using the rank biased overlap (RBO) measure to bring out the difference in the feature rankings between the two methods. RBO measures the set overlap of top selected elements in two lists. We run SHAP-TE and TI on the entire test set and obtain a list of FAs corresponding to each test point. We then compute the RBO corresponding to the ranked feature list of each test data point. 
Since the top ranked features are more relevant in explaining the output, we report RBO by considering different number of top elements from each ranked list.
Table \ref{tab:rbo} shows the median RBO values for different values of $k$. As observed, SHAP-TE and TI begin to differ significantly in rankings if we consider Top-$3$ and Top-$5$ ranked features. A median value of $0.61$ implies that for more than $50\%$ of the data points, only $1$ feature out of top $3$ is the same for these methods. This motivates us to investigate the differences in their behaviour in more detail, as described below.
\begin{table}
    \parbox{.45\linewidth}{
    \centering
    \caption{Median Rank biased Overlap values of SHAP-TE and TI.} 
    \label{tab:rbo}
     \begin{tabular}{|c|c| } 
         \hline
         $\mathbf{k}$ & \textbf{RBO}  \\ 
         \hline
         All & $0.77$ \\
         \hline
         Top-$5$ & $0.65$ \\
         \hline
         Top-$3$ & $0.61$ \\
         \hline
    \end{tabular}
    }
    \hfill
    \parbox{.45\linewidth}{
    \centering
    \caption{Median Variance of attribution values on test set.}
    \begin{tabular}{|c|c|c|}
    \hline
        $\mathbf{k}$ & \textbf{SHAP-TE} & \textbf{TI} \\
         \hline
        All & $4.9 \times 10^{-4}$ & $6.8 \times 10^{-4}$ \\
        Top-$5$ & $5.4 \times 10^{-4}$ & $7.8 \times 10^{-4}$ \\
        Top-$3$ & $3.6 \times 10^{-4}$ & $5.6 \times 10^{-4}$ \\
        \hline
    \end{tabular}
    \label{tab:Variance}
    }
\end{table}

\subsection{Significance of Attribution Ranking}
Since we are using the FA values to generate an ordered ranking of features, it is necessary to note the the amount of variation in attribution values to interpret the significance of obtained orderings. Higher variance in attribution values would imply a more significant ordering of features.\footnote{For eg, consider $2$ lists of attribution values $S_1=[1, 1.1, 1.3]$ and $S_2=[1, 3, 5]$. The ranking obtained from values in $S_2$ is more reliable than $S_1$.} For each data point, we measure the variance in the magnitude of Top $k$ FA values and report the median for complete data set. 
From Table \ref{tab:Variance}, we observe that TI captures nearly $50\%$ more variance in the attribution values than SHAP-TE. This implies that rankings obtained from TI are more significant than from SHAP-TE.

\subsection{Attribution Accuracy: How correctly are the right features attributed}



Figure \ref{fig:FlowChart} describes the outline of experiment to evaluate the correctness of feature attribution. To establish feature association, we apply
SHAP-TE and TI to the trained Random Forest model to produce FA values of each feature for every test data point. These values represent the contribution of each feature in producing an output. If we have two data points which are almost same, except for changes across a few feature values, we expect to see change in the output to be attributed to these differing features. As proposed by Griffon \cite{Shao_2019}, contribution of features for deviation in outputs can be computed by taking a difference of FA values.

\subsubsection{Implicit Interventional Measure}
\label{interventional_measure_discussion}
In this scenario, we consider all the data points where only one of the treatment variable differs and all other variables including covariates remain the same. We believe this subset of data essentially emulates the
real world scenario where one system parameter is controlled and remaining query variables are the same.

From the results in Table \ref{tab:interventional_acc}, we see that TI outperforms SHAP-TE for $2$ out of $3$ Treatment variables for both Top-$1$ and Top-$3$ measures. It is worth noting that the performance of SHAP-TE improves significantly from Top-$1$ to Top-$3$. This is because SHAP-TE always prefers \textit{IndexLevel} over the other two treatment variables 
However, \textit{PageCost} or \textit{MemoryLevel} do appear in the attributions at Rank $2$ or $3$. 

\begin{table}
    \parbox{.45\linewidth}{
    \centering
    \caption{Implicit Interventional Attribution Accuracy (S-TE stands for SHAP-TE).}
    \begin{tabular}{|c|c|c|c|c|}
        \hline
        Treatment & \multicolumn{2}{|c}{Top-1} & \multicolumn{2}{|c|}{Top-3} \\
        \hline 
        \hline
        & S-TE & TI & S-TE &
        TI \\
        \hline
        IndexLevel & 85\% & 68\% & 94\% & 91\% \\
        PageCost & 46\%  & 82\% & 84\% & 99\% \\
        MemoryLevel & 62\% & 79\% & 86\%  & 98\% \\
        \hline
        \textbf{Average} & \textbf{64}\% & \textbf{76}\% & \textbf{88}\% & \textbf{96}\% \\
        \hline 
    \end{tabular}
    \label{tab:interventional_acc}
    }
    \hfill
    \parbox{.45\linewidth}{
    \centering
    \caption{Explicit Intervention Attribution Accuracy (S-TE stands for SHAP-TE).}
    \centering
    \begin{tabular}{|c|c|c|c|c|}
        \hline
        Treatment & \multicolumn{2}{|c}{Top-1} & \multicolumn{2}{|c|}{Top-3} \\
        \hline 
        \hline
        & S-TE & TI & S-TE &
        TI \\
        \hline
        IndexLevel & 81\% & 80\% & 92\% & 97\% \\
        PageCost &  55\% & 82\% & 83\% & 96\% \\
        MemoryLevel & 46\% & 83\% & 84\% & 97\% \\
        \hline
        \textbf{Average} & \textbf{61}\% & \textbf{82}\% & \textbf{86}\% & \textbf{97}\% \\
        \hline 
    \end{tabular}
    \label{tab:explicit_intervention}
    }
\end{table}

\subsubsection{Explicit Interventional Measure}
\label{explicit_interventional_measure_discussion}
This measure enables us to use the whole test set for evaluating performance for each treatment variable. We compute the explicit interventional attribution accuracy of both FAMs on the whole test set. Table \ref{tab:interventional_acc} shows attribution accuracy measure when the true feature was among the top-$1$ or top-$3$ attributed features. We observe that average attribution accuracy of TI is better than that of SHAP-TE in both cases. Even for individual treatment variables, TI is significantly better in all cases except for Top-$1$ in \textit{IndexLevel}. 

\section{Conclusion}

We evaluated two prominent feature attribution methods for explaining tree-based models. The results show that the two methods differ in various aspects of efficacy and efficiency. In our case study, we observe that the amount of time that SHAP-TE takes to compute attribution values is nearly $\textbf{60}\mathbf{\times}$ higher than that of TI. This could be a potential constraint in certain large-scale computing applications. 

We also compared the performance accuracy of these methods using two different interventional approaches and observe that, on average, TI outperforms SHAP-TE. Based on these results, we conclude that despite the consistency guarantees, SHAP-TE does not provide benefits, in terms of attribution accuracy, in our case study of explaining job anomalies in cloud-computing applications. In addition, we have found that using TI provides high quality results at a lower computational footprint. 


We invite the research community to build on our findings. The code used for obtaining these results is available publicly\footnote{\href{https://github.com/sharmapulkit/TreeInterpretability\_AnomalyExplanation}{https://github.com/sharmapulkit/TreeInterpretability\_AnomalyExplanation}}.

\section*{Acknowledgements}

We thank our mentors, Javier Burroni and Prof. Andrew McCallum, for their guidance. We also thank Minsoo Thigpen for organizational support, as well as Scott Lundberg for providing insightful suggestions on a earlier draft. Finally, we thank anonymous reviewers for their feedback.



%
%
%
\bibliographystyle{acm}
\bibliography{TreePaper}

\begin{thebibliography}{10}

\bibitem{empericalRF}
{\sc Caruana, R., Karampatziakis, N., and Yessenalina, A.}
\newblock An empirical evaluation of supervised learning in high dimensions.
\newblock In {\em Proceedings of the 25th International Conference on Machine
  Learning\/} (2008), ICML ’08, p.~96–103.

\bibitem{anomDetSurvey}
{\sc Chandola, V., Banerjee, A., and Kumar, V.}
\newblock Anomaly detection: A survey.
\newblock {\em ACM Comput. Surv. 41\/} (07 2009).

\bibitem{runtimeanomaly}
{\sc {Cuzzocrea}, A., {Mumolo}, E., and {Cecolin}, R.}
\newblock Runtime anomaly detection in embedded systems by binary tracing and
  hidden markov models.
\newblock In {\em 2015 IEEE 39th Annual Computer Software and Applications
  Conference\/} (2015), vol.~2, pp.~15--22.

\bibitem{anotherRFExample}
{\sc Duque~Anton, S., Sinha, S., and Schotten, H.}
\newblock Anomaly-based intrusion detection in industrial data with svm and
  random forests.
\newblock pp.~1--6.

\bibitem{NIPS2019_9345}
{\sc Gentzel, A., Garant, D., and Jensen, D.}
\newblock The case for evaluating causal models using interventional measures
  and empirical data.
\newblock In {\em Advances in Neural Information Processing Systems 32},
  H.~Wallach, H.~Larochelle, A.~Beygelzimer, F.~d\textquotesingle
  Alch\'{e}-Buc, E.~Fox, and R.~Garnett, Eds. Curran Associates, Inc., 2019,
  pp.~11722--11732.

\bibitem{kuhn1953contributions}
{\sc Kuhn, H.~W., and Tucker, A.~W.}
\newblock {\em Contributions to the Theory of Games}, vol.~2.
\newblock Princeton University Press, 1953.

\bibitem{shapleyRegression}
{\sc Lipovetsky, S., and Conklin, M.}
\newblock Analysis of regression in game theory approach.
\newblock {\em Applied Stochastic Models in Business and Industry 17\/} (10
  2001), 319 -- 330.

\bibitem{treeExplainer}
{\sc Lundberg, S.~M., Erion, G., Chen, H., DeGrave, A., Prutkin, J.~M., Nair,
  B., Katz, R., Himmelfarb, J., Bansal, N., and Lee, S.-I.}
\newblock From local explanations to global understanding with explainable ai
  for trees.
\newblock {\em Nature Machine Intelligence 2}, 1 (2020), 2522--5839.

\bibitem{shap}
{\sc Lundberg, S.~M., and Lee, S.-I.}
\newblock A unified approach to interpreting model predictions.
\newblock In {\em Advances in Neural Information Processing Systems 30}. 2017.

\bibitem{padmulti}
{\sc {Peiris}, M., {Hill}, J.~H., {Thelin}, J., {Bykov}, S., {Kliot}, G., and
  {Konig}, C.}
\newblock Pad: Performance anomaly detection in multi-server distributed
  systems.
\newblock In {\em 2014 IEEE 7th International Conference on Cloud Computing\/}
  (2014), pp.~769--776.

\bibitem{otherRFExample}
{\sc {Primartha}, R., and {Tama}, B.~A.}
\newblock Anomaly detection using random forest: A performance revisited.
\newblock In {\em 2017 International Conference on Data and Software
  Engineering (ICoDSE)\/} (2017), pp.~1--6.

\bibitem{lime}
{\sc Ribeiro, M.~T., Singh, S., and Guestrin, C.}
\newblock "why should {I} trust you?": Explaining the predictions of any
  classifier.
\newblock In {\em Proceedings of the 22nd {ACM} {SIGKDD} International
  Conference on Knowledge Discovery and Data Mining, San Francisco, CA, USA,
  August 13-17, 2016\/} (2016), pp.~1135--1144.

\bibitem{treeInterpreterGit}
{\sc Saabas, A.}
\newblock Treeinterpreter.
\newblock \url{https://github.com/andosa/treeinterpreter}.

\bibitem{Shao_2019}
{\sc Shao, L., Srinivasan, S., Curino, C., Karanasos, K., Zhu, Y., Liu, S.,
  Eswaran, A., Lieber, K., Mahajan, J., Thigpen, M., and et~al.}
\newblock Griffon.
\newblock {\em Proceedings of the ACM Symposium on Cloud Computing - SoCC
  ’19\/} (2019).

\bibitem{deeplift17}
{\sc Shrikumar, A., Greenside, P., and Kundaje, A.}
\newblock Learning important features through propagating activation
  differences.
\newblock {\em CoRR abs/1704.02685\/} (2017).

\bibitem{deeplift16}
{\sc Shrikumar, A., Greenside, P., Shcherbina, A., and Kundaje, A.}
\newblock Not just a black box: Learning important features through propagating
  activation differences.
\newblock {\em CoRR abs/1605.01713\/} (2016).

\bibitem{shapleySampling}
{\sc {\v S}trumbelj, E., and Kononenko, I.}
\newblock Explaining prediction models and individual predictions with feature
  contributions.
\newblock {\em Knowledge and Information Systems\/} (2013).

\bibitem{SultaniCVPR}
{\sc Sultani, W., Chen, C., and Shah, M.}
\newblock Real-world anomaly detection in surveillance videos.
\newblock In {\em The IEEE Conference on Computer Vision and Pattern
  Recognition}.

\bibitem{rbo}
{\sc Webber, W., Moffat, A., and Zobel, J.}
\newblock A similarity measure for indefinite rankings.
\newblock {\em ACM Trans. Inf. Syst. 28}, 4 (Nov. 2010).

\bibitem{anomalyEEG}
{\sc {Wulsin}, D., {Blanco}, J., {Mani}, R., and {Litt}, B.}
\newblock Semi-supervised anomaly detection for eeg waveforms using deep belief
  nets.
\newblock In {\em 2010 Ninth International Conference on Machine Learning and
  Applications\/} (2010), pp.~436--441.

\end{thebibliography}
\end{document}